\documentclass[letterpaper, 10 pt, conference]{ieeeconf}  

\IEEEoverridecommandlockouts                              

\usepackage{amsmath,amssymb,graphicx}
\usepackage{float, booktabs}
\usepackage{multirow}
\usepackage{pifont}

\usepackage[
backend=biber,
sorting=none
]{biblatex}
\title{\LARGE \bf
Region-Affinity Attention for Whole-Slide Breast Cancer Classification in Deep Ultraviolet Imaging
}
\author{Nagur Shareef Shaik$^{1*}$ Teja Krishna Cherukuri$^{1*}$, Dong Hye Ye$^{1\dag}$ \\ $^1$Department of Computer Science, Georgia State University, Atlanta, GA USA
\thanks{$^{*}$Equal Contribution}
\thanks{$^{\dag}$Corresponding Author: \tt\small dongye@gsu.edu}
}

\begin{document}

\maketitle
\thispagestyle{empty}
\pagestyle{empty}

\begin{abstract}
Breast cancer diagnosis demands rapid and precise tools, yet traditional histopathological methods often fall short in intra-operative settings. Deep Ultraviolet (DUV) fluorescence imaging emerges as a transformative approach, offering high-contrast, label-free visualization of whole-slide images (WSIs) with unprecedented detail, surpassing conventional hematoxylin and eosin (H\&E) staining in speed and resolution. However, existing deep learning methods for breast cancer classification, predominantly patch-based, fragment spatial context and incur significant preprocessing overhead, limiting their clinical utility. Moreover, standard attention mechanisms, such as Spatial, Squeeze-and-Excitation, Global Context and Guided Context Gating, fail to fully exploit the rich, multi-scale regional relationships inherent in DUV-WSI data, often prioritizing generic feature recalibration over diagnostic specificity. This study introduces a novel Region-Affinity Attention mechanism tailored for DUV-WSI breast cancer classification, processing entire slides without patching to preserve spatial integrity. By modeling local neighbor distances and constructing a full affinity matrix, our method dynamically highlights diagnostically relevant regions, augmented by a contrastive loss to enhance feature discriminability. Evaluated on a dataset of 136 DUV-WSI samples, our approach achieves an accuracy of 92.67 $\pm$ 0.73\% and an AUC of 95.97\%, outperforming existing attention methods.
\end{abstract}


\section{Introduction}
\label{sec:intro}

Breast cancer, with an estimated 2.3 million new cases and 685,000 deaths annually, remains the most prevalent cancer among women worldwide, underscoring the urgent need for rapid and accurate diagnostic tools \cite{sung2021global}. Its heterogeneity, encompassing subtypes such as invasive ductal carcinoma (IDC), invasive lobular carcinoma (ILC), and ductal carcinoma in situ (DCIS), presents unique diagnostic challenges due to varying morphological features like nuclear pleomorphism, mitotic activity, and ductal structures \cite{zhang2025unique}. Early and precise diagnosis is critical for tailoring treatment strategies, particularly in intraoperative settings like breast-conserving surgery, where real-time margin assessment can prevent recurrence \cite{pradipta2020emerging}. Traditional histopathological methods, reliant on hematoxylin and eosin (H\&E) staining and microscopic examination, remain the gold standard but are hindered by time-intensive sample preparation, inter-observer variability, and dependence on expert pathologists \cite{cirecsan2013mitosis}. 

These limitations are particularly pronounced in intraoperative scenarios, where delays can compromise surgical outcomes. Deep Ultraviolet (DUV) fluorescence Whole-Slide Imaging (WSI) offers a transformative solution, providing label-free, high-contrast visualization of breast tissue morphology with superior resolution and speed compared to H\&E staining \cite{to2022deep, shamai2022deep}. By capturing intrinsic fluorescence signals from cellular components like tryptophan and collagen, DUV-WSI highlights critical diagnostic features without chemical staining, reducing processing time and enabling real-time applications \cite{ghahfarokhi2024deep}. However, the computational analysis of DUV-WSI for automated breast cancer classification faces challenges in preserving spatial coherence across large-scale images and dynamically prioritizing diagnostically relevant regions, such as tumor margins and cellular atypia.

Current deep learning approaches for WSI-based breast cancer classification predominantly rely on patch-based processing, where high-resolution images are segmented into smaller patches for feature extraction and classification \cite{to2022deep, ghahfarokhi2024deep}. While this strategy mitigates computational constraints, it fragments spatial relationships, disrupting the contextual integrity essential for accurate diagnosis \cite{veta2014breast}. To address this, attention mechanisms have been integrated into deep learning pipelines to selectively focus on informative regions. For instance, Spatial Attention reweights pixel-wise importance to highlight salient areas \cite{zhu2019empirical}, while Squeeze-and-Excitation (SE) enhances channel-wise feature dependencies to improve feature discriminability \cite{hu2018squeeze}. Similarly, Global Context Attention aggregates features across the entire image to capture broader dependencies \cite{cao2020global}, and Guided Context Gating (GCG) refines information flow by modulating feature interactions \cite{cherukuri2024guided}. However, these mechanisms fall short in fully exploiting the rich, multi-scale structural relationships inherent in DUV-WSI data. 

Spatial Attention struggles to capture long-range dependencies, SE prioritizes global feature recalibration over localized diagnostic relevance, Global Context Attention risks diluting critical local features, and GCG lacks explicit modeling of local neighborhood relationships \cite{cao2020global, cherukuri2024guided}. Furthermore, the reliance on patch-based processing introduces significant preprocessing overhead and fails to leverage the full spatial integrity of WSIs, limiting the clinical utility of these approaches for real-time intraoperative diagnostics. The limitations of existing methods underscore the need for a novel computational framework that preserves spatial coherence, dynamically emphasizes diagnostically relevant regions, and ensures computational efficiency for DUV-WSI analysis. 

To address these gaps, we propose the Region-Affinity Attention (RAA) mechanism, a novel attention approach designed specifically for breast cancer classification from DUV-WSI images. RAA processes entire slides, without patching, preserving spatial integrity while dynamically modeling region-specific affinities via a local neighborhood matrix. This design allows the network to prioritize diagnostically important regions, such as tumor boundaries and cellular atypia. Unlike conventional patch-based or transformer-style attention frameworks, RAA maintains spatial coherence and is optimized for computational efficiency in high-resolution settings. To further enhance discriminability, we incorporate a contrastive learning objective, which separates morphologically subtle classes in feature space. Together, these innovations enable a scalable, interpretable pipeline suitable for real-time, intraoperative applications in digital pathology. Our core contributions are:
\begin{itemize}
    \item We introduce a novel Region-Affinity Attention (RAA) mechanism that models local neighborhood relationships using a learnable affinity matrix, enabling context-aware feature refinement tailored to DUV-WSI.
    \item We propose a dual-objective training strategy that combines cross-entropy and contrastive supervision to improve the model’s ability to differentiate morphologically similar tissue classes in high-resolution whole-slide images.
\end{itemize}

\section{Related Work}
\label{sec:related_work}

The field of computational pathology has increasingly leveraged attention mechanisms to overcome the limitations of patch-based deep learning on Whole-Slide Images (WSIs). While numerous attention modules, such as Spatial Attention \cite{zhu2019empirical}, Squeeze-and-Excitation \cite{hu2018squeeze}, Global Context \cite{cao2020global}, and Guided Context Gating \cite{cherukuri2024guided}, have been adopted in medical imaging \cite{cherukuri2025dynamic, shaik2024spatial}, they generally operate on isolated feature recalibration principles and do not explicitly model the spatial affinity between local regions. This becomes particularly limiting for high-resolution fluorescence modalities like DUV-WSI, where diagnostically relevant regions are defined by nuanced structural relationships that span short spatial ranges.

More recent transformer-based approaches, such as Vision Transformers (ViT) \cite{dosovitskiy2021an}, offer global context modeling but often require aggressive patching or tokenization strategies due to their quadratic complexity \cite{afshin2025breast}. This not only reintroduces spatial fragmentation but also makes them less practical for intraoperative or real-time analysis of gigapixel-scale WSIs.

Graph-based and affinity-aware attention mechanisms have been explored in medical segmentation tasks \cite{wang2024label}, where local similarity between pixel neighborhoods is used to enhance spatial coherence. However, these techniques are rarely extended to classification pipelines and typically lack discriminative supervision (contrastive objectives) necessary for fine-grained tissue differentiation \cite{chaitanya2020contrastive}.

Region-Affinity Attention (RAA) advances this line of research by integrating learnable spatial affinity modeling with whole-slide, patch-free processing and supervised contrastive learning. Unlike prior attention methods, RAA constructs a localized affinity matrix over feature embeddings to preserve neighborhood structure and amplify diagnostically coherent regions. This is coupled with a dual-loss strategy that improves class separation in the embedding space, marking a novel contribution in the context of DUV-WSI classification.

\section{Methodology}

\begin{figure*}[t]
\centering
\includegraphics[width=\textwidth]{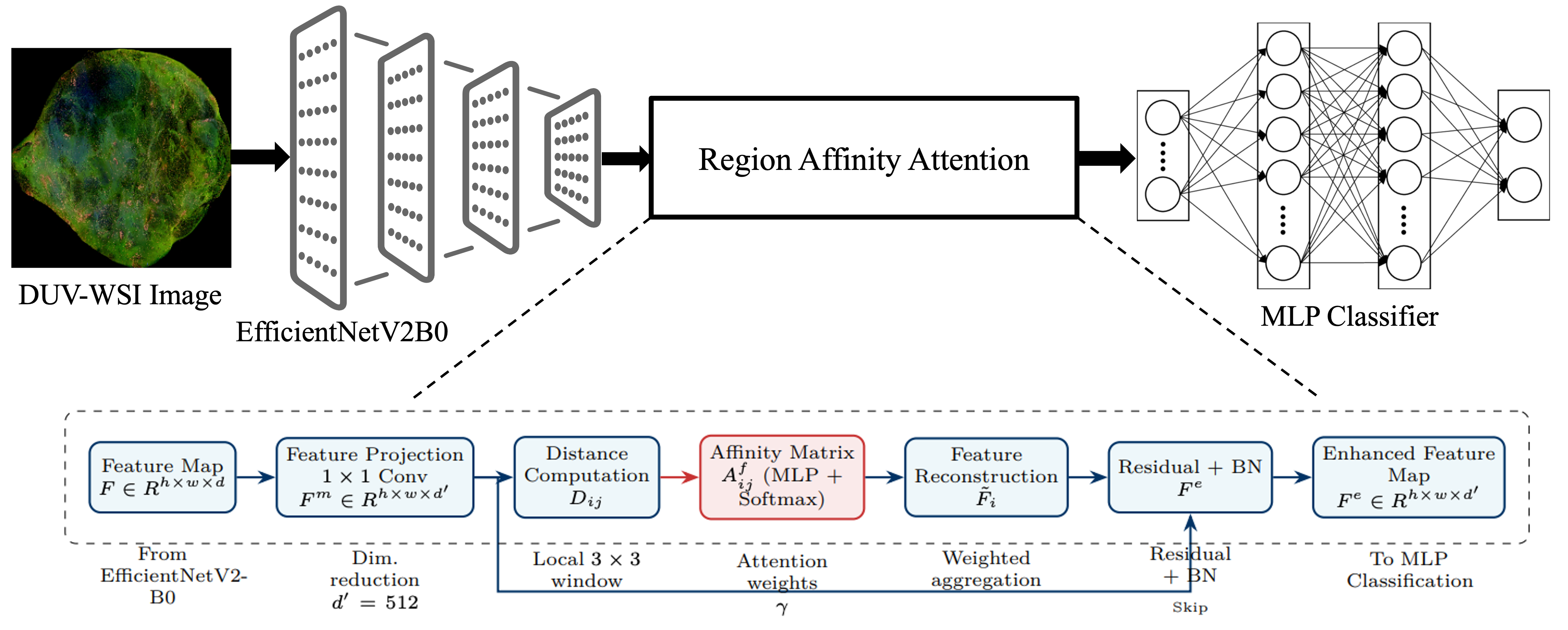}
\caption{Architecture of the Region-Affinity Attention (RAA) framework for breast cancer classification from DUV-WSI images. A pre-trained EfficientNetV2-B0 extracts spatial features, followed by the RAA module, which computes an affinity matrix based on local neighbor distances to prioritize diagnostically relevant regions. The refined features are classified using an MLP, with a combined cross-entropy and contrastive loss for optimization.}
\label{fig:raa_architecture}
\end{figure*}

In this research, we present a novel deep learning based approach for breast cancer classification from Deep Ultraviolet (DUV) fluorescence Whole-Slide Images (WSI). This methodology consists of three primary components: Spatial Representation Learning, Region-Affinity Attention Mechanism, and Classification with a Multi-Layer Perceptron (MLP). The pipeline is designed to process entire WSIs without patching, preserving spatial coherence and emphasizing diagnostically relevant regions. 

\subsection{Spatial Representation Learning}

To capture robust spatial features from high-resolution DUV-WSI images, we employ a pre-trained EfficientNetV2-B0 \cite{tan2021efficientnetv2}, a convolutional neural network (CNN) optimized for efficiency and accuracy through compound scaling. Pre-training on large-scale datasets, such as ImageNet \cite{deng2009imagenet}, enables the model to learn generalized feature representations, which are fine-tuned for the specific task of DUV-WSI breast cancer classification.

Given a DUV-WSI image $ X \in \mathbb{R}^{H \times W \times C} $, where $ H $ and $ W $ denote the height and width, and $ C = 3 $ represents the RGB channels, we resize the image to a fixed resolution of $ 256 \times 256 $ to balance computational efficiency and feature retention. The resized image is normalized to zero mean and unit variance to stabilize training. The EfficientNetV2-B0 model, parameterized by $ \theta_E $, extracts a feature map $ F $:
\begin{equation}
    F = \mathcal{E}(X; \theta_E), \quad F \in \mathbb{R}^{h \times w \times d},
    \label{eq:1}
\end{equation}
where $ h = w = 8 $ (after spatial reduction through convolutional layers) and $ d = 1280 $ is the feature dimension. The feature map $ F $ encodes multi-scale spatial information, capturing both local morphological patterns (cellular atypia) and global tissue structures critical for breast cancer diagnosis. Fine-tuning is performed by updating $ \theta_E $ on the DUV-WSI dataset, ensuring adaptation to the unique fluorescence properties of the images.

\subsection{Region-Affinity Attention}

The Region-Affinity Attention (RAA) is designed based on \cite{wang2024label}, to dynamically prioritize diagnostically relevant regions in the feature map $ F $, addressing the limitations of existing attention mechanisms that fail to capture multi-scale, region-specific relationships in DUV-WSI data. Unlike Spatial Attention \cite{zhu2019empirical} or Squeeze-and-Excitation \cite{hu2018squeeze}, which focus on pixel-wise or channel-wise recalibration, RAA explicitly models local neighborhood relationships through an affinity matrix, ensuring both local and global contextual awareness.

\subsubsection{Feature Projection}
The feature map $ F \in \mathbb{R}^{h \times w \times d} $ is first projected using a 1x1 convolutional layer to reduce dimensionality and enhance computational efficiency:
\begin{equation}
    F^m = \mathcal{P}(F; W_p, b_p), \quad F^m \in \mathbb{R}^{h \times w \times d'},
    \label{eq:2}
\end{equation}
where $ \mathcal{P} $ denotes the projection operation with learnable parameters $ W_p \in \mathbb{R}^{d \times d'} $ and $ b_p \in \mathbb{R}^{d'} $, and $ d' = 512 $ is the reduced feature dimension. This projection preserves essential feature information while reducing computational overhead.

\subsubsection{Affinity Matrix Construction}
For each pixel $ i \in \{1, 2, \dots, h \times w\} $ in the feature map $ F^m $, we compute pairwise distances to its neighboring pixels $ j \in \mathcal{K}(i) $, where $ \mathcal{K}(i) $ is a local neighborhood (e.g., a 3x3 window). The distance $ D_{ij} $ quantifies feature similarity between pixels $ i $ and $ j $:
\begin{equation}
    D_{ij} = \frac{1}{d'} \sum_{c=1}^{d'} | F^m_{i,c} - F^m_{j,c} |,
    \label{e:3}
\end{equation}
where $ F^m_{i,c} $ and $ F^m_{j,c} $ are the feature values at pixels $ i $ and $ j $ for channel $ c $. The L$_1$ distance ensures robustness to outliers in feature distributions, which is critical for DUV-WSI images with heterogeneous fluorescence patterns. 

The distance matrix $ D_{ij} $ is transformed through a two-layer Multi-Layer Perceptron (MLP) to capture subtle variations in local similarity modeling complex relationships:
\begin{equation}
    D'_{ij} = \text{ReLU}(\text{GELU}(\mathcal{M}_1(D_{ij}; W_1, b_1))),
    \label{eq:4}
\end{equation}
where $ \mathcal{M}_1 $ is the first MLP layer with parameters $ W_1 $ and $ b_1 $, and GELU (Gaussian Error Linear Unit) introduces non-linearity to capture intricate feature interactions \cite{hendrycks2016gaussian}. This step enables the network to learn an adaptive weighting function over pairwise distances rather than relying on a fixed or manually designed similarity kernel. The combination of GELU and ReLU activations balances smooth gradient flow and sparsity, allowing the model to modulate affinities with both fine-grained sensitivity and non-linearity. Such learned transformations are especially useful in DUV-WSI data, where fluorescence variability can distort raw distance metrics and make simple distance functions ineffective. The transformed distances $ D'_{ij} $ are then normalized to compute the affinity matrix $ A^f_{ij} $:
\begin{equation}
    A^f_{ij} = \frac{\exp(-\gamma D'_{ij})}{\sum_{j' \in \mathcal{K}(i)} \exp(-\gamma D'_{ij'})},
    \label{eq:5}
\end{equation}
where $ \gamma > 0 $ is a learnable scaling parameter that controls the sharpness of the affinity distribution. This softmax normalization ensures that $ \sum_{j \in \mathcal{K}(i)} A^f_{ij} = 1 $, assigning higher weights to neighbors with similar features, which correspond to diagnostically relevant regions (e.g., tumor clusters).

\subsubsection{Feature Reconstruction}
The affinity matrix $ A^f_{ij} $ is used to reconstruct the feature map, emphasizing regions with high diagnostic relevance:
\begin{equation}
    \tilde{F}_i = \sum_{j \in \mathcal{K}(i)} A^f_{ij} F^m_j,
    \label{eq:6}
\end{equation}
where $ \tilde{F}_i \in \mathbb{R}^{d'} $ is the reconstructed feature vector for pixel $ i $. This weighted aggregation enhances features from similar regions, effectively focusing on areas like tumor boundaries while suppressing background noise.

To stabilize training and promote feature reuse, a residual connection and batch normalization are applied:
\begin{equation}
    F^e = \text{ReLU}(\text{BN}(\tilde{F} + F^m)),
    \label{eq:7}
\end{equation}
where $ F^e \in \mathbb{R}^{h \times w \times d'} $ is the enhanced feature map, and BN denotes batch normalization \cite{bjorck2018understanding}. The residual connection ensures that the model retains original features while incorporating region-specific refinements.

\subsubsection{Significance}
The RAA mechanism draws on graph-based attention principles, where the affinity matrix $ A^f_{ij} $ resembles an adjacency matrix in a graph representation of the feature map \cite{velivckovic2017graph}. Unlike transformer-based attention \cite{vaswani2017attention}, which computes global pairwise interactions with quadratic complexity $ O((h \times w)^2) $, RAA restricts attention to local neighborhoods, reducing complexity to $ O(h \times w \times |\mathcal{K}|) $, where $ |\mathcal{K}| $ is the neighborhood size (e.g., 9 for a 3x3 window). This makes RAA scalable for large-scale WSI processing while maintaining robust feature modeling.

\subsection{Classification with MLP}
The enhanced feature map $ F^e $ is flattened into a one-dimensional vector $ \text{vec}(F^e) \in \mathbb{R}^{h \cdot w \cdot d'} $ and passed through a Multi-Layer Perceptron (MLP) for classification:
\begin{equation}
    Z = \phi(W_2 \cdot \text{ReLU}(W_1 \cdot \text{vec}(F^e) + b_1) + b_2),
    \label{eq:8}
\end{equation}

where $ W_1 \in \mathbb{R}^{(h \cdot w \cdot d') \times d_h} $, $ W_2 \in \mathbb{R}^{d_h \times 2} $, $ b_1 \in \mathbb{R}^{d_h} $, and $ b_2 \in \mathbb{R}^2 $ are learnable parameters, with $ d_h = 512 $ as the hidden dimension. The softmax activation $ \phi $ produces class probabilities for benign (0) and malignant (1) classes: $\hat{Y} = \arg\max Z$. This MLP maps the refined features to a binary classification output, leveraging the discriminative power of $ F^e $.

\subsection{Loss Function}
To optimize the model, we combine a cross-entropy loss with a contrastive loss to enhance feature discriminability. The cross-entropy loss $ L_{CE} $ is defined as:
\begin{equation}
    L_{CE} = -\frac{1}{N} \sum_{i=1}^N y_i \log(p_i),
    \label{eq:9}
\end{equation}
where $ N $ is the number of samples, $ y_i \in \{0, 1\} $ is the ground truth label, and $ p_i $ is the predicted probability for the true class. The contrastive loss $ L_{CL} $ encourages the model to learn separable feature representations by minimizing the distance between samples of the same class and maximizing the distance between samples of different classes:
\begin{equation}
\begin{aligned}
L_{CL} &= \frac{1}{N} \sum_{i=1}^N \sum_{j \neq i} \Bigg[ y_i y_j \max(0, \|f_i - f_j\|_2 - m_1)^2 \\
&\quad + (1 - y_i y_j) \max(0, m_2 - \|f_i - f_j\|_2)^2 \Bigg],
\end{aligned}
\end{equation}
where $ f_i = \text{vec}(F^e_i) $ is the flattened feature map for sample $ i $, $ ||\cdot||_2 $ is the L2 norm, $ m_1 = 0.5 $ is the margin for same-class samples, and $ m_2 = 2.0 $ is the margin for different-class samples. This contrastive objective encourages the model to pull together feature representations of samples from the same class while pushing apart those from different classes. Unlike traditional cross-entropy which only supervises the final output, the contrastive loss imposes structure in the latent space, which is especially beneficial for subtle morphological differences in DUV-WSI images. Our formulation is inspired by supervised contrastive learning \cite{khosla2020supervised}, adapted here to operate directly on affinity-refined whole-slide embeddings.

The total loss is $L = L_{CE} + \lambda L_{CL}$, where $ \lambda = 0.1 $ is a weighting factor determined via hyperparameter tuning. This combined loss ensures both accurate classification and robust feature separation, critical for distinguishing subtle morphological differences in DUV-WSI images.

\begin{table*}[!ht]
\centering
\caption{Ablation Study: evaluation of image encoders, RAA components, neighborhood sizes, input resolutions, and validation strategies. Metrics are reported as percentages, with mean $\pm$ standard deviation for cross-validation or single values for hold-out validation. Default configuration ( EfficientNetV2B0, 3$\times$3 neighborhood, 256$\times$256 resolution, 5-fold cross-validation)}
\vspace{-5pt}
\label{tab:comprehensive_ablation}
\small
\begin{tabular}{llcccc}
\toprule
\textbf{Configuration} & \textbf{Value} & \textbf{Accuracy (\%)} & \textbf{F$_1$-Score (\%)} & \textbf{Quadratic Kappa (\%)} & \textbf{AUC (\%)} \\
\midrule
\multirow{3}{*}{\textbf{Image Encoder}} & DenseNet121 & 90.45 $\pm$ 0.83 & 90.11 $\pm$ 1.06 & 80.30 $\pm$ 1.36 & 95.00 $\pm$ 0.90 \\
& ResNet50 & 92.65 $\pm$ 0.68 & 92.39 $\pm$ 0.90 & 84.81 $\pm$ 1.85 & 95.33 $\pm$ 1.25 \\
& Vision Transformer & 90.60 $\pm$ 1.0 & 90.25 $\pm$ 1.20 & 80.29 $\pm$ 1.31 & 94.60 $\pm$ 1.37 \\
\midrule
\multirow{7}{*}{\textbf{RAA}} & w/o Contrastive Loss & 89.00 $\pm$ 1.16 & 88.69 $\pm$ 1.45 & 77.60 $\pm$ 1.37 & 93.06 $\pm$ 1.33 \\
& w/o MLP & 90.42 $\pm$ 0.81 & 90.16 $\pm$ 0.87 & 80.42 $\pm$ 1.46 & 92.78 $\pm$ 1.18 \\
& ReLU-ReLU MLP & 92.67 $\pm$ 1.58 & 92.53 $\pm$ 1.70 & 85.07 $\pm$ 1.16 & 95.22 $\pm$ 1.46 \\
& GELU-GELU MLP & 88.94 $\pm$ 0.66 & 88.74 $\pm$ 0.81 & 77.53 $\pm$ 1.30 & 93.07 $\pm$ 1.68 \\
& $\mathcal{K}$ (5$\times$5) & 91.91 $\pm$ 0.61 & 91.68 $\pm$ 0.74 & 83.42 $\pm$ 1.53 & 92.95 $\pm$ 1.38 \\
& $\mathcal{K}$ (7$\times$7) & 90.45 $\pm$ 1.60 & 90.05 $\pm$ 1.79 & 80.27 $\pm$ 1.76 & 92.49 $\pm$ 2.12 \\
& $\mathcal{K}$ (9$\times$9) & 90.08 $\pm$ 1.26 & 89.72 $\pm$ 1.34 & 78.80 $\pm$ 2.65 & 93.87 $\pm$ 1.21 \\
\midrule
\textbf{Image Size} & 512$\times$512 & 91.22 $\pm$ 1.22 & 90.99 $\pm$ 1.34 & 82.00 $\pm$ 2.60 & 94.74 $\pm$ 1.61 \\
\midrule
\multirow{1}{*}{\textbf{Validation}} & 10-Fold & 92.69 $\pm$ 1.45 & 92.33 $\pm$ 1.26 & 84.79 $\pm$ 1.31 & 95.15 $\pm$ 1.21 \\
\midrule
\multicolumn{2}{l}{\textbf{Region-Affinity Attention (RAA)}} & \textbf{92.67 $\pm$ 0.73} & \textbf{92.36 $\pm$ 0.72} & \textbf{84.79 $\pm$ 0.96} & \textbf{95.97 $\pm$ 0.92} \\
\bottomrule
\end{tabular}
\end{table*}

\begin{table*}[!ht]
\centering
\caption{Performance Comparison of Attention Mechanisms on DUV-WSI Breast Cancer Classification (5-Fold Cross-Validation)}
\vspace{-5pt}
\begin{tabular}{lcccccc}
\toprule
\textbf{Method} & \textbf{Accuracy (\%)} & \textbf{Precision (\%)} & \textbf{Recall (\%)} & \textbf{F$_1$-Score (\%)} & \textbf{Quadratic Kappa(\%)} & \textbf{AUC(\%)} \\
\midrule
\text{EB2V0 (Baseline)} & 91.22 $\pm$ 0.84 & 91.18 $\pm$ 0.85 & 91.09 $\pm$ 0.80 & 90.98 $\pm$ 0.82 & 82.00 $\pm$ 1.10 & 92.71 $\pm$ 1.20 \\
\text{Spatial Attention} & 91.90 $\pm$ 0.79 & 91.85 $\pm$ 0.81 & 91.64 $\pm$ 0.74 & 91.72 $\pm$ 0.76 & 83.43 $\pm$ 1.05 & 93.51 $\pm$ 1.08 \\
\text{Squeeze \& Excitation} & 91.50 $\pm$ 0.83 & 91.90 $\pm$ 0.88 & 91.48 $\pm$ 0.81 & 91.32 $\pm$ 0.85 & 82.75 $\pm$ 1.15 & 93.13 $\pm$ 1.10 \\
\text{Global Context} & 91.28 $\pm$ 0.92 & 91.51 $\pm$ 0.76 & 91.31 $\pm$ 0.87 & 91.36 $\pm$ 0.84 & 82.37 $\pm$ 1.12 & 93.75 $\pm$ 1.24 \\
\text{Guided Context Gating} & 92.14 $\pm$ 0.92 & 89.97 $\pm$ 0.92 & 92.85 $\pm$ 0.79 & 92.57 $\pm$ 0.81 & 83.23 $\pm$ 1.10 & 93.91 $\pm$ 1.15 \\
\text{Region-Affinity Attention (RAA)} & \textbf{92.67 $\pm$ 0.73} & \textbf{93.32 $\pm$ 0.77} & \textbf{92.06 $\pm$ 0.70} & \textbf{92.36 $\pm$ 0.72} & \textbf{84.79 $\pm$ 0.96} & \textbf{95.97 $\pm$ 0.92} \\
\bottomrule
\end{tabular}
\label{table:attention_comparison}
\end{table*}

\section{Experimental Results}

\subsection{Dataset}
The dataset used in this study consists of 136 Deep Ultraviolet Fluorescence Whole-Slide Images (DUV-WSI) of breast cancer tissue samples, collected from pathology slides \cite{ghahfarokhi2024deep}. Each image is large-scale, containing intricate pixel-level details that are crucial for distinguishing between benign and malignant tissues. The images were annotated with labels indicating whether the tumor present in each image was benign or malignant.

\subsection{Experimental Setup}
In this study, we employed a 5-fold cross-validation strategy to rigorously evaluate the performance of our proposed Region-Affinity Attention (RAA) mechanism for breast cancer classification from DUV-WSI images. The model was initialized with pre-trained weights to leverage transfer learning and improve performance. For training, we used the Adam optimizer with an initial learning rate of 0.001. Additionally, model checkpointing was implemented to save the best-performing model after each epoch based on validation accuracy. To assess the model's performance, we computed several evaluation metrics including accuracy, precision, recall, F1-score, quadratic kappa score, and AUC (Area Under the ROC Curve). These metrics provide a comprehensive evaluation of the model’s classification ability, with particular emphasis on both the balance between precision and recall, as well as the model's discriminatory power. All models were trained for a set number of epochs, and the best model for each fold was selected for final evaluation. Experiments were run on a single NVIDIA P100 GPU using PyTorch with a fixed seed of 42 for reproducibility. Hyperparameters were selected by grid search over learning rate, batch size, and contrastive weight $\lambda$, with $10^{-3}$, $16$, and $0.1$ chosen by validation accuracy. Source code will be released upon publication.

\subsection{Quantitative Evaluation}

\subsubsection{Ablation Study}

We conducted an extensive ablation study to systematically evaluate the architectural design of the Region-Affinity Attention (RAA) framework. This involved varying five key components: image encoder choice, module configuration, affinity neighborhood size, input resolution, and validation strategy. As shown in Table~\ref{tab:comprehensive_ablation}, among the tested encoders, EfficientNetV2B0 achieved the highest accuracy of $92.67\% \pm 0.73$, outperforming DenseNet121 ($90.45\%$,\,\textdownarrow~2.22) and Vision Transformer (ViT) ($90.60\%$,\,\textdownarrow~2.07). While ResNet50 performed comparably ($92.65\%$), EfficientNetV2B0 was selected as the primary encoder due to its favorable trade-off between computational efficiency and feature expressiveness, particularly important for high-resolution Deep Ultraviolet Whole-Slide Imaging (DUV-WSI). Notably, the choice of a CNN-based encoder like EfficientNetV2B0 supports faster inference and lower memory overhead compared to ViT, making it more suitable for real-time intra-operative margin assessment. Statistical analysis confirms that RAA with EfficientNetV2B0 significantly outperforms DenseNet121 and Vision Transformer (p = 0.043), while ResNet50 (p = 0.686) is comparable in 5-fold cross-validation Accuracy. These results underscore the importance of selecting an encoder architecture that captures both local morphology and broader spatial context without introducing excessive overhead.

Further analysis of individual RAA components revealed their distinct contributions to overall performance. Removing the contrastive loss led to a significant drop in accuracy from $92.67\%$ to $89.00\%$ (\textdownarrow~3.67), indicating its central role in encouraging discriminative feature learning across spatial regions. Likewise, eliminating the MLP component reduced accuracy to $90.42\%$ (\textdownarrow~2.25), suggesting that non-linear transformation of affinity-weighted features is essential for effective classification. To assess the impact of activation functions, we compared ReLU-ReLU and GELU-GELU configurations within the MLP. While ReLU preserved the original performance, GELU reduced accuracy to $88.94\%$ (\textdownarrow~3.73), likely due to its smoother non-linearity being less effective in small-sample settings. These observations reinforce the architectural decisions made in RAA, particularly the inclusion of contrastive supervision and ReLU activations to support generalization.

The spatial scope of region affinity was explored by adjusting the neighborhood size in the affinity matrix. Expanding from the default $3\times3$ to $5\times5$ offered a modest improvement ($91.91\%$,\,\textuparrow~1.46), while larger neighborhoods of $7\times7$ and $9\times9$ degraded performance to $90.45\%$ and $90.08\%$, respectively (\textdownarrow~2.22, \textdownarrow~2.59), likely due to the inclusion of irrelevant or noisy context. We also evaluated the influence of image resolution, where the default $256\times256$ configuration was compared against $512\times512$. Despite the larger input size, performance declined to $91.22\%$ (\textdownarrow~1.45), indicating that diagnostically relevant features such as nuclear atypia are preserved at lower resolutions. The $256\times256$ setting thus offers a favorable trade-off between accuracy and computational efficiency, as the $512\times512$ configuration significantly increases computational load without corresponding performance gains. This balance is especially critical for real-time intra-operative margin assessment in clinical workflows. Across validation strategies, 5-fold and 10-fold cross-validation yielded consistent accuracies of $92.67\%$ and $92.69\%$. These findings confirm that RAA generalizes well under limited data conditions and that each component contributes meaningfully to its robust performance in high-resolution breast cancer classification.

\begin{figure*}[!ht]
    \centering
    \includegraphics[width=\textwidth]{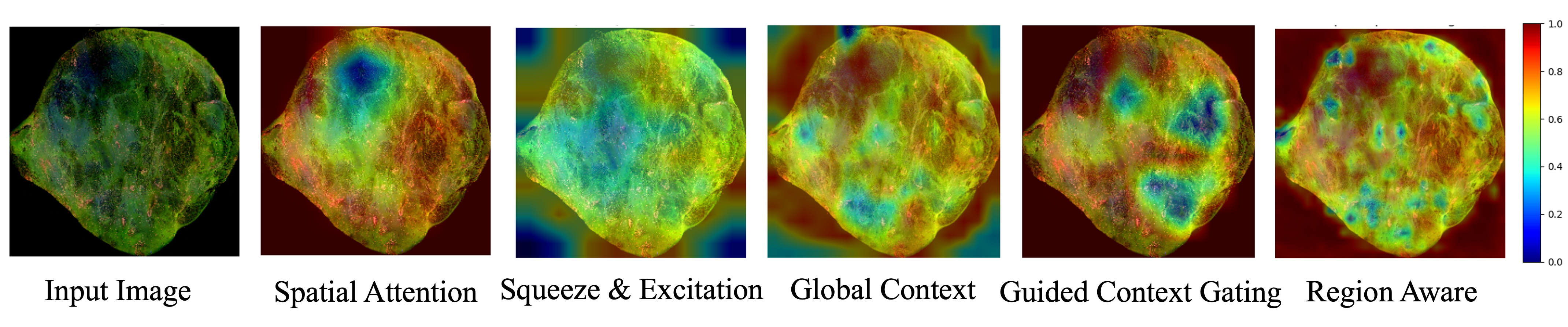}
    \vspace{-10pt}
    \caption{Qualitative evaluation of attention mechanisms using Grad-CAM++ visualizations for breast cancer classification from DUV-WSI images; The color bar indicates activation intensity, with warmer colors highlighting regions of high model attention, thus providing visual interpretability of the attention mechanism.}
    \label{fig:attention_maps}
\end{figure*}

\subsubsection{Comparison Study}
The results in Table \ref{table:attention_comparison} demonstrate that the proposed Region-Affinity Attention (RAA) method outperforms existing attention mechanisms in breast cancer classification from DUV-WSI images. RAA achieves the highest accuracy (92.67\% $\pm$ 0.73), precision (93.32\% $\pm$ 0.77), recall (92.06\% $\pm$ 0.70), F1-score (92.36\% $\pm$ 0.72), and AUC (95.97\% $\pm$ 0.92), highlighting its superior ability to correctly identify cancerous regions while minimizing false positives and negatives. In contrast, methods like Spatial Attention, Squeeze \& Excitation, and Guided Context Gating show lower performance, with RAA’s F1-score and AUC standing out as key indicators of its improved discriminative power.

The significant improvement in RAA’s performance can be attributed to its novel design, which dynamically captures multi-scale, region-specific relationships through its attention mechanism, unlike traditional methods that focus on global or spatial recalibration. Although RAA achieved the highest accuracy among all attention mechanisms, its improvement over Guided Context Gating was not statistically significant ($p = 0.268$), highlighting its consistent performance even in closely matched scenarios. This focus on locally relevant areas enhances both classification accuracy and model interpretability. However, RAA’s higher computational complexity and reliance on pre-trained EfficientNetV2-B0 for feature extraction are potential limitations. Despite these, RAA’s ability to outperform existing methods makes it a promising solution for real-time, whole-slide breast cancer diagnosis using DUV-WSI images. Benchmarking against transformer-only and Multiple Instance Learning (MIL) pipelines (e.g., CLAM, TransMIL) is outside this study's attention-focused scope; the ViT row in Table~\ref{tab:comprehensive_ablation} provides an initial transformer reference, with broader comparisons left for future work.

\subsection{Qualitative Evaluation}

The attention maps generated using Grad-CAM++ \cite{chattopadhay2018grad} shown in Figure~\ref{fig:attention_maps} highlight the differences in feature importance across various attention mechanisms. The Region-Affinity Attention (RAA) method demonstrates superior focus on diagnostically relevant regions, effectively localizing high-importance areas while suppressing less informative background noise. In contrast, Spatial Attention and Squeeze \& Excitation show broader activation patterns, often diluting focus across non-critical regions. Global Context Attention exhibits dispersed activations, reducing localized interpretability, while Guided Context Gating produces slightly improved localization but still lacks the fine-grained adaptability of RAA. The enhanced region-specific attention in RAA enables better tumor region identification, making the model more interpretable and reliable for clinical decision-making in breast cancer classification. Grad-CAM++ is used rather than visualizing the affinity matrix $A^f_{ij}$ directly, so that all mechanisms in Figure~\ref{fig:attention_maps} are compared under a uniform, architecture-agnostic framework---baselines such as Squeeze-and-Excitation and Global Context do not expose spatially resolved maps for direct visualization.

\section{Conclusion}

In this study, we introduced Region-Affinity Attention (RAA) for breast cancer classification from DUV-WSI images, addressing the limitations of existing attention mechanisms. By leveraging spatial feature extraction via EfficientNetV2-B0, followed by region-sensitive affinity computation, RAA dynamically enhances diagnostically relevant regions, improving both classification accuracy and interpretability. Experimental results demonstrated that RAA outperforms traditional attention methods, achieving the highest accuracy (92.67\%), macro F1-score (92.36\%), and ROC AUC (95.97\%), affirming its effectiveness in precise tumor region localization. Qualitative evaluation using Grad-CAM++ further validated the superior focus of RAA on key pathological areas, reinforcing its explainability. While RAA effectively models local context, it does not capture long-range dependencies beyond fixed neighborhoods. Future work could address this limitation by exploring hierarchical or multi-scale affinity mechanisms. Although our study focuses on label-free DUV-WSI, the proposed framework is modality-agnostic and may be extended to H\&E-stained images to enhance its generalizability. Evaluation is also limited to 136 WSIs from a single source; multi-center cohorts and comparisons with transformer-based and MIL pipelines will be needed for broader generalizability. RAA is intended as a decision-support tool for pathologists, not an autonomous diagnostic system.

\printbibliography[title={References}]

\end{document}